\def\year{2022}\relax
\documentclass[letterpaper]{article} % DO NOT CHANGE THIS
\usepackage{aaai22}  % DO NOT CHANGE THIS
\usepackage{times}  % DO NOT CHANGE THIS
\usepackage{helvet}  % DO NOT CHANGE THIS
\usepackage{courier}  % DO NOT CHANGE THIS
\usepackage[hyphens]{url}  % DO NOT CHANGE THIS
\usepackage{graphicx} % DO NOT CHANGE THIS
\usepackage{natbib}  % DO NOT CHANGE THIS AND DO NOT ADD ANY OPTIONS TO IT
\usepackage{caption} % DO NOT CHANGE THIS AND DO NOT ADD ANY OPTIONS TO IT
\DeclareCaptionStyle{ruled}{labelfont=normalfont,labelsep=colon,strut=off} % DO NOT CHANGE THIS
\usepackage{algorithm}
\usepackage{algorithmic}

\usepackage{latexsym}
\usepackage{amsmath}
\usepackage{amssymb}
\usepackage{mathtools}
\usepackage{xcolor}
\usepackage{booktabs}
\usepackage{multirow}
\usepackage{caption}

\usepackage{hyperref}
\hypersetup{
    colorlinks=true,
    citecolor=black,
    urlcolor=black,
}

\usepackage{newfloat}
\usepackage{listings}
\floatstyle{ruled}
\newfloat{listing}{tb}{lst}{}
\floatname{listing}{Listing}
\title{\textsc{ValueNet}: A New Dataset for Human Value Driven Dialogue System}
\author {
    Liang Qiu\textsuperscript{\rm 1},
    Yizhou Zhao\textsuperscript{\rm 1},
    Jinchao Li\textsuperscript{\rm 2},
    Pan Lu\textsuperscript{\rm 1},
    Baolin Peng\textsuperscript{\rm 2}, \\
    Jianfeng Gao\textsuperscript{\rm 2},
    Song-Chun Zhu\textsuperscript{\rm 1}
}
\affiliations {
    \textsuperscript{\rm 1}UCLA Center for Vision, Cognition, Learning, and Autonomy \\
    \textsuperscript{\rm 2}Microsoft Research, Redmond \\
    liangqiu@ucla.edu
}
\begin{document}

\maketitle

\begin{abstract}
Building a socially intelligent agent involves many challenges, one of which is to teach the agent to speak guided by its value like a human. However, value-driven chatbots are still understudied in the area of dialogue systems. Most existing datasets focus on commonsense reasoning or social norm modeling. In this work, we present a new large-scale human value dataset called \textsc{ValueNet}, which contains human attitudes on 21,374 text scenarios. The dataset is organized in ten dimensions that conform to the basic human value theory in intercultural research. We further develop a Transformer-based value regression model on \textsc{ValueNet} to learn the utility distribution. 
Comprehensive empirical results show that the learned value model could benefit a wide range of dialogue tasks. For example, by teaching a generative agent with reinforcement learning and the rewards from the value model, our method attains state-of-the-art performance on the personalized dialog generation dataset: \textsc{Persona-Chat}. With values as additional features, existing emotion recognition models enable capturing rich human emotions in the context, which further improves the empathetic response generation performance in the \textsc{EmpatheticDialogues} dataset. To the best of our knowledge, \textsc{ValueNet} is the first large-scale text dataset for human value modeling, and we are the first one trying to incorporate a value model into emotionally intelligent dialogue systems. The dataset is available at \url{https://liang-qiu.github.io/ValueNet/}.

% \pan{summary and cons of existing datasets}
% Towards this end, we propose to incorporate a value model into dialogue agents.
% \pan{motivations, key statistics, features of our dataset.}
% \pan{models and experiments.}
% \pan{Summary.}

% Case studies show that the value model provides a new way for numerical user profiling.

% \textcolor{orange}{[TO CHECK]: rewrite the contribution, the three tasks} Experimental results show that the connotations of value information help improve the performance of dialogue generation, raise the precision of emotion prediction, and increase the accuracy of predicting characters' behavior in a fantasy-story world. 
% It also shows that the framework could predict user's value priority from their dialog history.
\end{abstract}

%------------------------------------------- introduction -------------------------------------------------
\section{Introduction}
Value refers to desirable goals in human life. They guide the selection or evaluation of actions, policies, people, and events. A person's value priority or hierarchy profoundly affects his or her attitudes, beliefs, and traits, making it one core component of personality~\citep{schwartz2012overview}. In dialogue systems, modeling human values is a critical step towards building socially intelligent chatbots~\citep{qiu2021towards}. By considering values, we can estimate user behavior and cognitive patterns from their utterances and generate responses that conform to the robot's persona configuration. For example, the robot is set to be aware of human values, and it invites Jerry to drink beers, but Jerry replies, ``\textit{You know that is tempting but is not good for our fitness}". The bot could read from the dialogue that Jerry prefers a healthy and self-disciplined lifestyle and steer its recommendation to healthier options in the future. 
\begin{figure}[t]
    \centering
    \includegraphics[width = \linewidth]{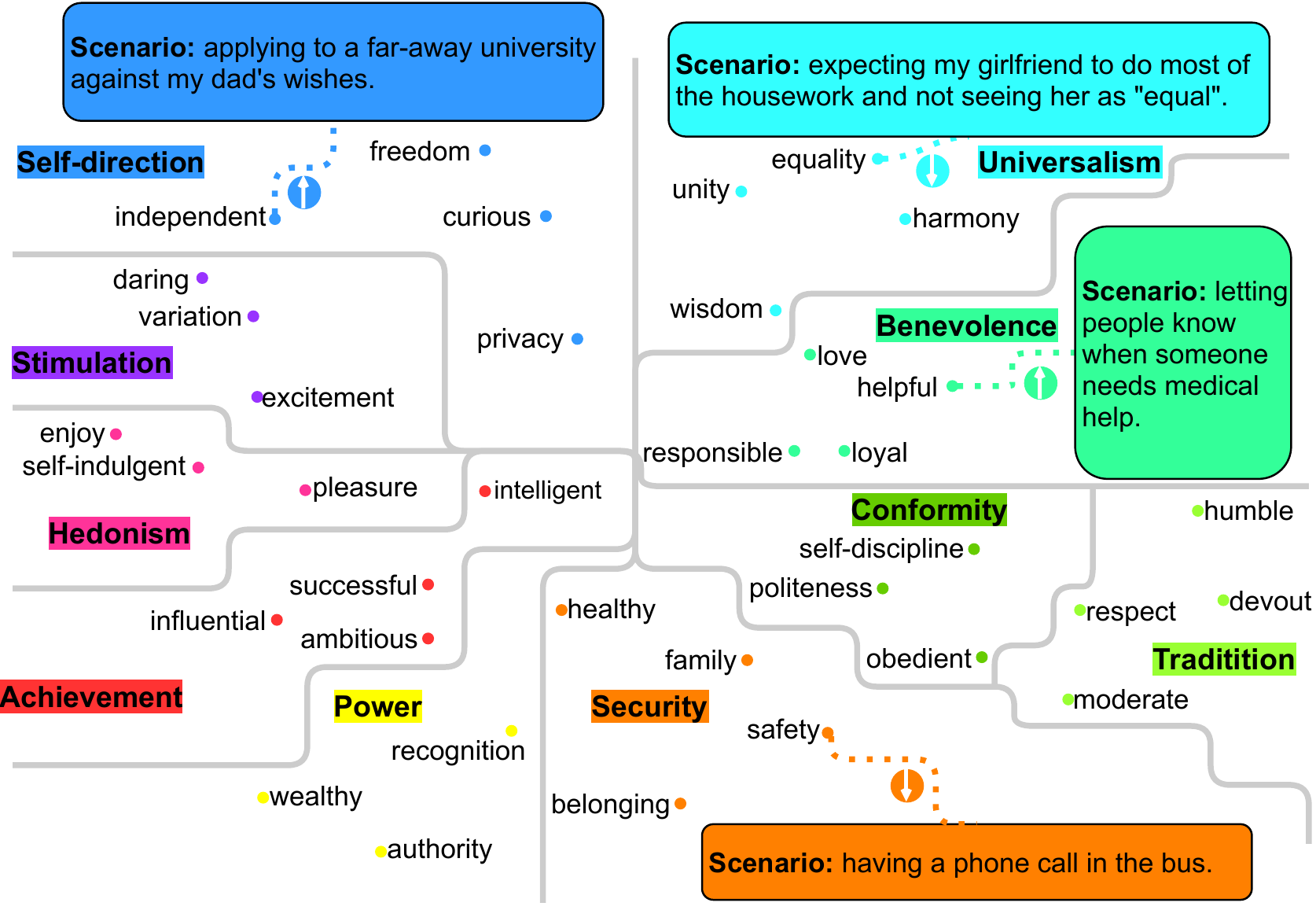}
    \caption{The presented \textsc{ValueNet} dataset with curated social scenarios organized by Schwartz values~\citep{schwartz2012overview}.}
    \label{fig:schwartz_values}
\end{figure}

The development of socially intelligent chatbots has been one of the longest-running goals in artificial intelligence. Early dialogue systems such as Eliza~\citep{weizenbaum1966eliza}, Parry~\citep{colby1971artificial}, and more recent SimSimi\footnote{\url{https://simsimi.com/}}, Panda Ichiro~\citep{okuda2018ai}, Replika~\citep{fedorenko2018avoiding}, XiaoIce~\citep{zhou-etal-2020-design}, were designed to mimic human behavior and incorporate emotional quotients (EQ) to some extent. There are also datasets and benchmarks for studying related problems, such as emotion recognition~\citep{mckeown2010semaine, hsu-etal-2018-emotionlines, poria-etal-2019-meld, ghosal-etal-2020-cosmic}, personalized dialogue generation~\citep{zhang-etal-2018-personalizing, liu-etal-2020-impress}, and empathetic dialogue generation~\citep{rashkin-etal-2019-towards}. Even though value plays a fundamental and critical role in human EQ, there is a lack of explicit modeling of values in the dialogue domain, based on social domain theory. We have seen recent efforts about crowdsourcing social commonsense knowledge base or benchmarks~\citep{forbes-etal-2020-social, sap-etal-2019-social, lourie2020scruples, hendrycks2020aligning,hwang2020comet,gabriel2020paragraph}. However, it is not clearly shown how an agent can leverage this knowledge to estimate the users' value priorities or guide its own speaking and actions. In this paper, we aim to alleviate this problem and investigate the usage of a learned value function.

% \pan{motivation, collection, statistics, and features of the dataset.}
We start the study by curating a knowledge base of human values called \textsc{ValueNet}. Samples with value-related scenarios were identified based on value-defined keyword searching. Next, we asked Amazon Mechanical Turk workers about how the provided scenarios will affect one's value. This is based on the assumption that values underlie our attitudes; they are the guideline by which we evaluate things. Workers assess behaviors/events positively if they promote or protect the attainment of the goals we value. Behaviors/events are evaluated negatively if they hinder or threaten the attainment of these valued goals. The whole process gives us a large-scale (over 21k samples) multi-dimensional knowledge base of value. Figure~\ref{fig:schwartz_values} shows the overall structure of \textsc{ValueNet}. Each split represents a value dimension identified in the theory of basic human values~\citep{schwartz2012overview}. The figure also illustrates the value-related keywords and scenarios. The circular arrangement of the values represents a motivational continuum. By organizing data in such a structure, we anticipate the \textsc{ValueNet} to provide comprehensive coverage of different aspects of human values.

% \pan{benchmarks, methods, experiments.}
Next, we develop a Transformer-based value model to evaluate the utility score suggesting the positive or negative judgment given an utterance. We provide a detailed analysis of learning with multiple Transformer variants. Then we conduct a wide range of experiments to demonstrate that the value model could benefit EQ-related dialogue tasks: \textit{(i)} By finetuning a generative agent with reinforcement learning and the reward from our value model, the method achieves state-of-the-art performance on the personalized dialogue dataset: \textsc{Persona-Chat}~\citep{zhang-etal-2018-personalizing}; \textit{(ii)} By incorporating values as additional features, in \textsc{EmpatheticDialogues}~\citep{rashkin-etal-2019-towards}, we improve the emotion classification accuracy of existing models, which further facilitates the empathetic response generation; \textit{(iii)} Visualization of the value model shows that it provides a numerical way of user profile modeling from their utterances.

% Finally, given the value function, we are allowed to estimate the user's value priority. 
% By further incorporating reinforcement learning into language model training, we achieve a new value-driven model for personality-consistent dialogue generation. 
% And by incorporating values as additional features, we improve the emotion classification accuracy based on the same Transformer classifier. 
% . \textcolor{red}{[TODO: the empirical results show that ...]}

% \pan{dataset, models, experiments}
In all, our contributions are two-fold. First, we present a large-scale dataset \textsc{ValueNet} for the modeling of human values that are well-defined in intercultural research.
Second, we initiate to develop the value model learned from \textsc{ValueNet} to several EQ-related tasks and demonstrate its usage for building a value-driven dialogue system.
% Second, we initiate to apply human values to estimate user profiles and generate personalized conversations. 
Our methodology can be generalized to a wide range of interactive situations in socially aware dialogue systems~\citep{zhao2018sogo}, and human-robot interactions~\citep{yuan2017development, liang2021human}. 

\section{Related Work}
An abundance of related work inspires our work. Our work aims to make contributions to dialogue systems by incorporating the theory of human value. The dataset we collect shares a similar nature with multiple social commonsense benchmarks and knowledge bases. Besides, we apply our \textsc{ValueNet} for various dialogue tasks related to EQ.

\subsection{Theory of Human Value and Utility} 
In the field of intercultural research,~\citet{schwartz2012overview} developed the theory of basic human values. The theory identifies ten basic personal values that are recognized across cultures and explains where they come from, as shown in Figure \ref{fig:schwartz_values}. The closer any two values in either direction around the circle, the more similar their underlying motivations are; the more distant, the more antagonistic their motivations. Note that dividing the value item domain into ten distinct values is an arbitrary convenience. It is reasonable to partition the value items into more or less fine-tuned distinct values according to the needs and objectives of one's analysis\footnote{A refinement of the theory~\citep{schwartz2012refining}, partitions the same continuum into $19$ more narrowly defined values that permit more precise explanation and prediction. We use the original $10$-dimension version for simplicity in this paper.}. Similarly, in the economics field, the concept of utility~\cite{fishburn1970utility} is initially defined as a measure of pleasure or satisfaction in economics and ethics that drives human activities at all levels. Therefore, when we teach agents to speak and act in a socially intelligent way, an approach considering human value utilities should be adopted. In this paper, we aim to learn a utility function for each dimension of value and steer the dialogue system response generation accordingly.

\subsection{Social Commonsense Benchmarks} 
\citet{hendrycks2020aligning} present the \textsc{ETHICS} dataset, a benchmark that assesses a language model's knowledge of basic concepts of morality.
% , including justice, well-being, duties, virtues, and commonsense morality.
\textsc{Scruples}~\citep{lourie2020scruples} is a large-scale dataset with ethical judgments over real-life anecdotes, motivated by descriptive ethics. \textsc{Social-Chem-101} presented by~\citet{forbes-etal-2020-social} is a corpus that catalogs rules-of-thumb as basic concept units for studying people's everyday social norms and moral judgments. They also propose Neural Norm Transformer to reason about previously unseen situations, generating relevant social rules-of-thumb. \textsc{Social IQA}~\citep{sap-etal-2019-social} is a large-scale benchmark for commonsense reasoning about social situations. 
% It could also be used as a resource for transfer learning of commonsense knowledge.
~\citet{he-etal-2017-learning} present a task and corpus for predicting the preferable options from two sentences describing the scenarios that may involve social and cultural situations. Instead, in this work, we release a new dataset \textsc{ValueNet} that provides annotation of human attitudes from different value aspects.

\subsection{Emotionally Intelligent Dialogue Datasets} 
% There are artificial intelligence companions like XiaoIce~\citep{zhou-etal-2020-design} taking account emotional quotients in its system design. XiaoIce dynamically recognizes human feelings and states, understands user intents, and responds to user needs to satisfy the human expectation for communication, affection, and social belonging. 
Several datasets are presented to study emotion dynamics in dialogues. DailyDialog~\cite{li-etal-2017-dailydialog} is a multi-turn dialogue dataset, which reflects the way of daily communication and provides emotion labels for speakers.~\citet{hsu-etal-2018-emotionlines} present EmotionLines with emotions labeling on all utterances in each dialogue based on their textual content. MELD~\citep{poria-etal-2019-meld} is an extension of EmotionLines for multi-modal multi-party emotion recognition. \citet{mckeown2010semaine} record a corpus SEMAINE of emotionally coloured conversations. \citet{ghosal-etal-2020-cosmic} propose a framework COSMIC for emotion recognition in conversations by considering mental states, events, actions, and cause-effect relations. 
DialogRE~\cite{yu-etal-2020-dialogue} is the first human-annotated dialogue-based dataset for social relation inference~\cite{qiu-etal-2021-socaog}.
% For personalized dialogue generation,\cite{li-etal-2016-persona}]}
\textsc{Persona-Chat}~\citep{zhang-etal-2018-personalizing} (revised in ConvAI2~\citep{dinan2020second}) provides natural language profiles of speakers. 
% They train conditional generation models based on the system and human personas. They also show preliminary results predicting profile information about the interlocutors. 
Based on \textsc{Persona-Chat}, ~\citet{liu-etal-2020-impress} propose a transmitter-receiver-based framework with explicitly human understanding modeling to enhance the quality of personalized dialogue generation.
\textsc{EmpatheticDialogues}~\citep{rashkin-etal-2019-towards} is a dataset that provides 25k conversations grounded in emotional situations. Each dialogue is grounded in a specific situation where a speaker was feeling a given emotion.

\section{The \textsc{ValueNet} Dataset}

During decision-making, people tend to pick the choice that aligns more with their own values. This work aims to provide a transferable knowledge base for human value modeling in natural language. To collect the \textsc{ValueNet} dataset, we curated social scenarios with value-related keywords and further annotated them via Amazon Mechanical Turk. Each sample in \textsc{ValueNet}  is a social scenario description labeled with the annotator's attitude through a specific value lens. 

The entire dataset is organized in a circular structure as shown in Figure~\ref{fig:schwartz_values}, aligning with the theory of basic human values~\cite{schwartz2012overview}. The theory identifies ten universal values that are recognized throughout major cultures. The circular structure reflects the dynamic relations among these values, \textit{i.e.,} the pursuit of some value may result in either accordance with another value or a conflict with another value. The ten distinct values can be further organized into four higher-order groups.
\begin{itemize}
  \item \textbf{Openness to change}: self-direction, stimulation
  \item \textbf{Self-enhancement}: hedonism, achievement, power
  \item \textbf{Conservation}: security, conformity, tradition
  \item \textbf{Self-transcendence}: benevolence, universalism
\end{itemize}
We describe the collection details of the \textsc{ValueNet} in the following sections.

\begin{figure}[t]
    \centering
    \includegraphics[width = 0.95\linewidth]{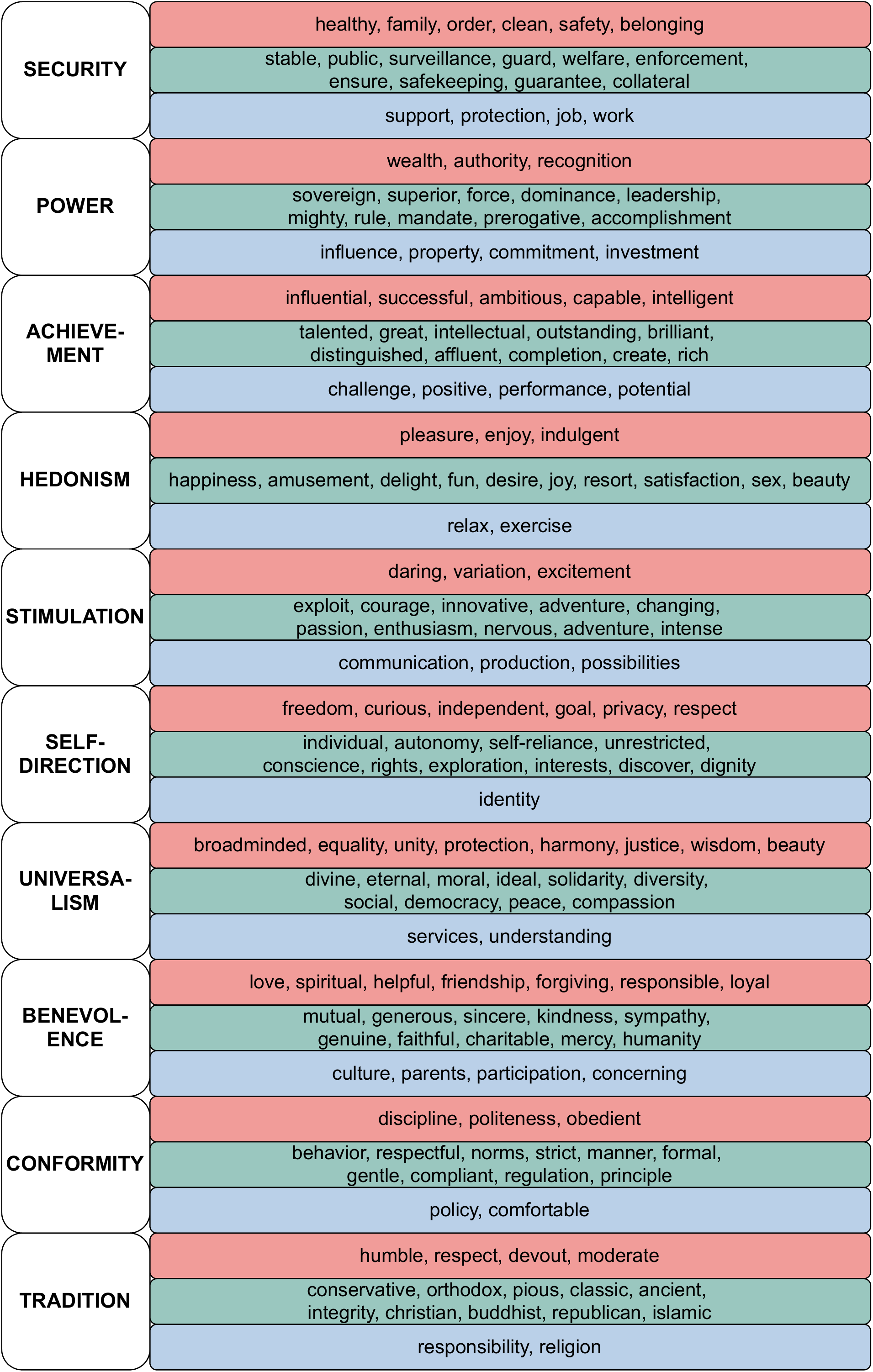}
    \caption{Ten universal human values and related keywords for social scenario curation. \textbf{Red}: keywords in the original value definition~\cite{schwartz2012overview}; \textbf{Green}: associated keywords found with datamuse; \textbf{Blue}: associated keywords found with GloVe embedding.}
    \label{fig:valuenet_keywords}
\end{figure}

\subsection{Social Scenario Curation} 
We curated a set of 21,374 social scenarios from the large-scale social-related database \textsc{Social-Chem-101}~\citep{forbes-etal-2020-social}.
%,  \textsc{Social IQa}~\citep{sap-etal-2019-social}, \textsc{OpenBookQA}~\citep{mihaylov-etal-2018-suit}, \textsc{CommonsenseQA}~\citep{talmor-etal-2019-commonsenseqa}, \textsc{ETHICS}~\citep{hendrycks2020aligning}. 
Value-related scenarios are retrieved with value keywords after lemmatization and stemming. There are three sets of keywords identified for each dimension of Schwartz value: (1) the keywords in the original definition of each value in Schwartz's paper~\cite{schwartz2012overview}; (2) words that share a similar meaning, words that are often used to describe the original keywords, and words that are triggered by (strongly associated with) the original keywords\footnote{We use datamuse (\url{https://www.datamuse.com/api/}) for this purpose.}; (3) words that are near the original keywords in the GloVe~\citep{pennington-etal-2014-glove} embedding space. The value keywords are verified and confirmed by humans as listed in Figure~\ref{fig:valuenet_keywords}.

\subsection{Value-Aspect Attitude Annotation} 
\begin{figure}[t]
    \centering
    \includegraphics[width = 0.9\linewidth]{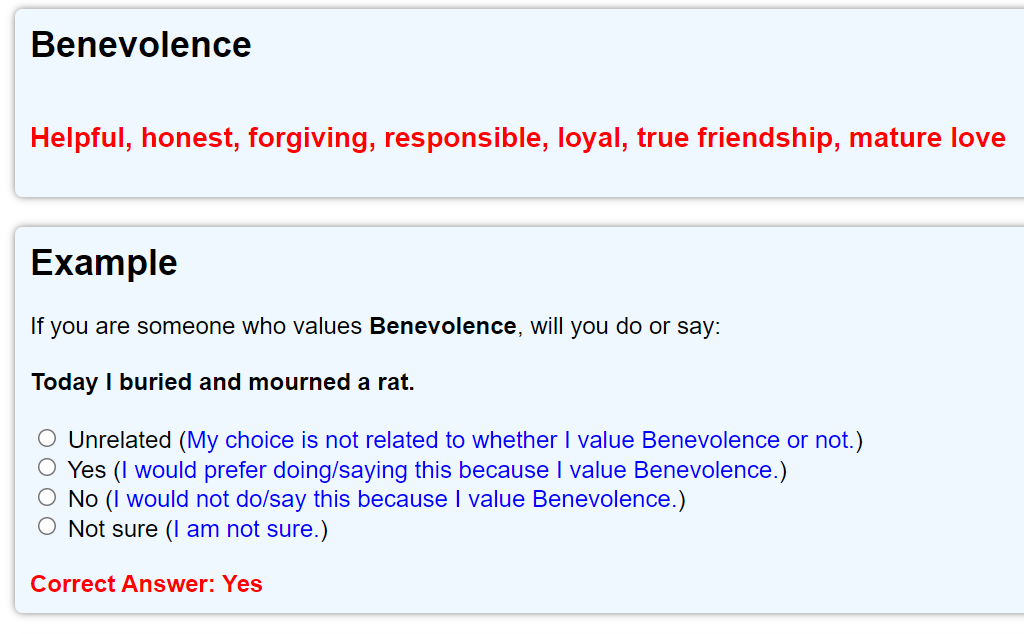}
    \caption{Value-aspect attitude annotation in AMT.}
    \label{fig:amt_example}
\end{figure}

We crowdsourced people's attitudes to the curated scenarios on Amazon Mechanical Turk (AMT). Figure~\ref{fig:amt_example} shows an example.

We follow a strict procedure to select qualified workers and ensure the workers understand the concept of each value we ask. In Figure~\ref{fig:amt_example}, the definition of \texttt{BENEVOLENCE} is shown to the workers throughout the entire annotation process. To further help the understanding, we include three examples in each assignment with correct answers being ``yes", ``no", and ``unrelated", respectively. The worker is then required to answer a prerequisite question correctly to proceed to the formal survey. The formal survey is composed of ten questions, including two hidden qualification checking questions. Before publishing on the AMT, two Ph.D. students prepared the qualification questions by annotating a small subset of the curated scenarios. Their agreed samples (100 in total) were randomly inserted into the survey for worker selection. The selection procedure was done in the value dimensions with more scenarios to get a large pool of qualified workers and a relatively balanced final dataset across different values. The complete Mechanical Turk interface is attached in the Appendix for reference.

\begin{figure}[th]
    \centering
    \includegraphics[width = 0.9\linewidth]{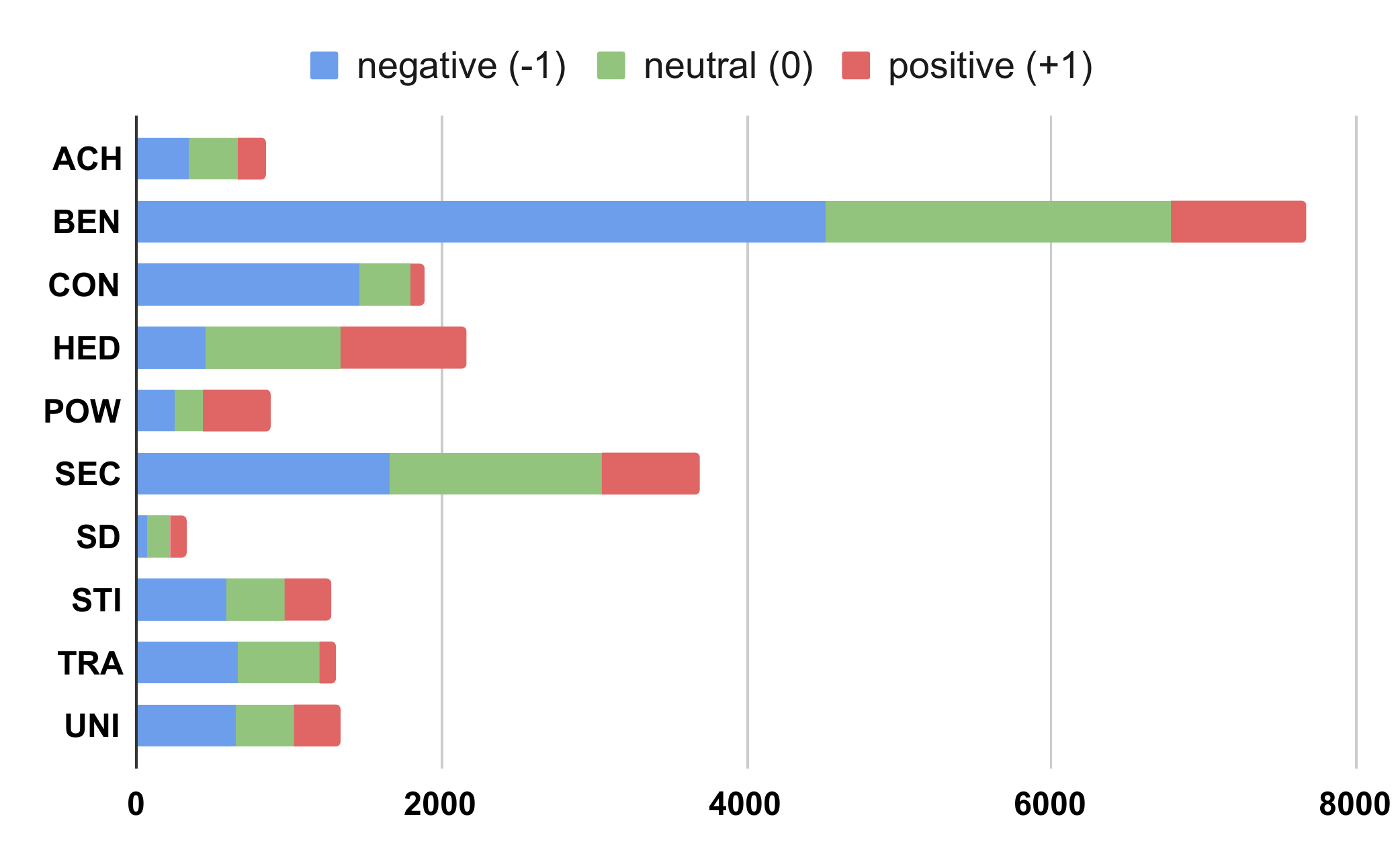}
    \caption{The sample number and label distribution of each value split in the \textsc{ValueNet}.}
    \label{fig:valuenet_stats}
\end{figure}

A total of 681 experienced AMT workers participated in our \textsc{ValueNet} annotation. 443 of them passed the qualification test. Each scenario is assigned to four different workers. The original inter-annotator agreement is 64.9\%, and the Fleiss' kappa score~\cite{fleiss1971measuring} among the workers is 0.48, which considers the possibility of the agreement by chance. Keeping the scope of \textsc{ValueNet} in commonly-agreed attitudes towards social scenarios, we only retain the samples with three or more agreements. Figure~\ref{fig:valuenet_stats} shows the sample size of each value split and their label distribution. 

% \begin{table}[ht]
% \centering
% \small
% \resizebox{0.6\linewidth}{!}{
% \begin{tabular}{llr}
% \toprule
% \multicolumn{3}{l}{\textsc{ValueNet} }               \\
% \midrule
% \multirow{4}{*}{\# samples}     & train   & 16,030  \\
%               & valid   & 3,206  \\
%               & test    & 2,138  \\
%               & total   & 21,374 \\
% \midrule
% \multirow{3}{*}{average \# tokens} & train  & 12.05   \\
%                  & valid  & 12.09   \\
%                  & test   & 12.26   \\
%                  & total  & 12.07   \\
% \midrule
% \multirow{3}{*}{unique \# tokens}  & train   & 12,452  \\
%                  & valid   & 5,292   \\
%                  & test    & 4,112   \\
%                  & total   & 14,143  \\
% \bottomrule
% \end{tabular}
% }
% \caption{Statistics of the \textsc{ValueNet} dataset.}
% \label{tab:valuenet_stats2}
% \end{table}

The data is split into the train (75\%), valid (15\%), and test (10\%). Similar to the polarity in sentiment analysis~\cite{kouloumpis2011twitter}, we quantify the annotated labels into numerical values: yes (positive): +1, no (negative): -1, unrelated (neutral): 0. We denote the numerical values as \textbf{utility} to describe the effect of a scenario on one's value. In other words, for people who appreciate a certain value, actions with a higher utility in this value dimension would be more desirable to them. 
\begin{table}[]
\centering
\resizebox{0.9\linewidth}{!}{
\begin{tabular}{ccccc}
\toprule
\textsc{ValueNet}          & train  & valid & test  & total  \\
\midrule
\# samples        & 16,030 & 3,206 & 2,138 & 21,374 \\
average \# tokens & 12.05  & 12.09 & 12.26 & 12.07  \\
unique \# tokens  & 12,452 & 5,292 & 4,112 & 14,143 \\
\bottomrule
\end{tabular}
}
\caption{Statistics of the \textsc{ValueNet} dataset.}
\label{tab:valuenet_stats2}
\end{table}

Table~\ref{tab:valuenet_stats2} shows more statistical details about the \textsc{ValueNet} dataset. In total, we collected 21,374 samples covering a wide range of scenarios in daily social life.

\begin{table*}[]
\centering
\small
\renewcommand\tabcolsep{5.0pt} 
\resizebox{0.9\linewidth}{!}{
\begin{tabular}{llccccccccccc}
\toprule
     &               & \multicolumn{1}{l}{$\mathbf{F}1$(-1)} & \multicolumn{1}{l}{$\mathbf{F}1$(0)} & \multicolumn{1}{l}{$\mathbf{F}1$(1)} & \multicolumn{1}{l}{$\mathbf{P}$(-1)} & \multicolumn{1}{l}{$\mathbf{P}$(0)} & \multicolumn{1}{l}{$\mathbf{P}$(1)} & \multicolumn{1}{l}{$\mathbf{R}$(-1)} & \multicolumn{1}{l}{$\mathbf{R}$(0)} & \multicolumn{1}{l}{$\mathbf{R}$(1)} & \multicolumn{1}{l}{\textbf{Acc.}$\uparrow$} & \multicolumn{1}{l}{\textbf{MSE}$\downarrow$} \\
\midrule
\multicolumn{1}{l|}{\multirow{5}{*}{\textsc{ValueNet} (original)}}  & \multicolumn{1}{l|}{fastText}   & 0.70      & 0.46   & 0.43   & 0.65    & 0.47   & 0.55   & \textbf{0.76}   & 0.44   & 0.35   & 0.58       & 0.66  \\
\multicolumn{1}{l|}{}  & \multicolumn{1}{l|}{BERT}       & \textbf{0.73}    & 0.50    & 0.51   & 0.72    & 0.46   & 0.71   & 0.74    & 0.55   & 0.39   & 0.61       & 0.39  \\
\multicolumn{1}{l|}{}  & \multicolumn{1}{l|}{DistilBERT} & 0.71     & 0.52   & 0.47   & 0.74    & 0.45   & 0.69   & 0.68    & 0.62   & 0.36   & 0.60       & \textbf{0.37} \\
\multicolumn{1}{l|}{}  & \multicolumn{1}{l|}{RoBERTa}    & 0.65     & 0.51   & 0.34   & 0.74    & 0.40   & 0.71   & 0.58    & 0.69   & 0.22   & 0.55       & 0.41  \\
\multicolumn{1}{l|}{}  & \multicolumn{1}{l|}{BART}       & 0.00     & \textbf{0.76}  & 0.54   & 0.00    & 0.70   & 0.60   & 0.00    & \textbf{0.83}  & \textbf{0.49}  & \textbf{0.67}      & 0.52  \\
\midrule
\multicolumn{1}{l|}{\multirow{5}{*}{\textsc{ValueNet} (balanced)}}  & \multicolumn{1}{l|}{fastText}   & 0.70     & 0.48   & 0.43   & 0.64    & 0.50   & 0.54   & \textbf{0.76}   & 0.45   & 0.36   & 0.59       & 0.68  \\
\multicolumn{1}{l|}{}  & \multicolumn{1}{l|}{BERT}       & 0.67     & 0.48   & 0.51   & 0.73    & 0.42   & 0.61   & 0.62    & 0.58   & 0.43   & 0.57       & 0.40   \\
\multicolumn{1}{l|}{}  & \multicolumn{1}{l|}{DistilBERT} & 0.66     & 0.49   & 0.50   & 0.74    & 0.41   & 0.61   & 0.60    & 0.60   & 0.43   & 0.57       & 0.40  \\
\multicolumn{1}{l|}{}  & \multicolumn{1}{l|}{RoBERTa}    & 0.65     & 0.51   & 0.34   & 0.74    & 0.40   & 0.71   & 0.58    & 0.69   & 0.22   & 0.55       & 0.41  \\
\multicolumn{1}{l|}{}  & \multicolumn{1}{l|}{BART}       & 0.00     & 0.75   & 0.51   & 0.00    & 0.72   & 0.57   & 0.00    & 0.77   & 0.47   & 0.65       & 0.55  \\
\midrule
\multicolumn{1}{l|}{\multirow{5}{*}{\textsc{ValueNet} (augmented)}} & \multicolumn{1}{l|}{fastText}   & 0.58     & 0.52   & 0.29   & 0.72    & 0.40    & 0.65   & 0.49    & 0.75   & 0.18   & 0.52       & 0.59  \\
\multicolumn{1}{l|}{}  & \multicolumn{1}{l|}{BERT}       & 0.67     & 0.55   & 0.41   & 0.78    & 0.43   & \textbf{0.78}  & 0.58    & 0.76   & 0.28   & 0.58       & 0.38  \\
\multicolumn{1}{l|}{}  & \multicolumn{1}{l|}{DistilBERT} & 0.68     & 0.57   & 0.41   & \textbf{0.79}   & 0.44   & \textbf{0.78}  & 0.59    & 0.78   & 0.28   & 0.60       & 0.38  \\
\multicolumn{1}{l|}{}  & \multicolumn{1}{l|}{RoBERTa}    & 0.70     & 0.56   & 0.41   & 0.78    & 0.45   & 0.75   & 0.64    & 0.74   & 0.28   & 0.61       & 0.40   \\
\multicolumn{1}{l|}{}  & \multicolumn{1}{l|}{BART}       & 0.00     & 0.74   & \textbf{0.57}  & 0.00    & \textbf{0.75}  & 0.49   & 0.00    & 0.73   & 0.66   & 0.64       & 0.46   \\
\bottomrule
\end{tabular}
}
\caption{Value modeling performance in the \textsc{ValueNet} dataset. \textbf{Bold} items are the best in each metric column.}
\label{tab:valuenet_learning}
\end{table*}

\begin{table*}[]
\centering
\small
\resizebox{0.85\linewidth}{!}{
\begin{tabular}{l|cccccccccc}
\toprule
\textbf{Acc.}       & ACH  & BEN  & CON  & HED  & POW  & SEC  & SD   & STI  & TRA  & UNI  \\
\midrule
\textsc{ValueNet} (original)  & \textbf{0.56} & \textbf{0.68} & 0.82 & \textbf{0.63} & 0.35 & \textbf{0.52} & 0.45 & \textbf{0.58} & 0.60  & \textbf{0.51} \\
\textsc{ValueNet} (balanced)  & 0.53 & 0.58 & \textbf{0.83} & \textbf{0.63} & \textbf{0.41} & 0.50  & 0.42 & 0.53 & 0.61 & 0.50  \\
\textsc{ValueNet} (augmented) & 0.48 & 0.66 & 0.82 & 0.58 & 0.33 & 0.47 & \textbf{0.48} & 0.49 & \textbf{0.64} & 0.42 \\
\bottomrule
\end{tabular}
}
\caption{Accuracies of the BERT~\cite{devlin2018bert} value model across different value dimensions in the \textsc{ValueNet} dataset.}
\label{tab:valuenet_learning2}
\end{table*}

\section{Value Modeling}
We experiment using Transformer-based pre-trained language models for modeling human values from the \textsc{ValueNet} dataset.

\subsection{Task Formalization}
Given a social scenario $s$, we wish to learn a value function that models the utility distribution of $s$ from the ten Schwartz value dimensions: $\mathbf{V}(s) = [V_{\texttt{SEC}}(s), V_{\texttt{POW}}(s), V_{\texttt{ACH}}(s), V_{\texttt{HED}}(s), V_{\texttt{STI}}(s), V_{\texttt{SD}}(s), V_{\texttt{UNI}}(s)$ $, V_{\texttt{BEN}}(s), V_{\texttt{CON}}(s), V_{\texttt{TRA}}(s)]$, where $V_{\texttt{\$VALUE}}(\cdot) \in [-1, 1]$ and $V_{\texttt{\$VALUE}}(\cdot) \in \mathbb{R}$. 

\subsection{Model}
Pre-trained language model variants: BERT~\cite{devlin2018bert}, RoBERTa~\cite{liu2019roberta}, DistilBERT~\cite{sanh2019distilbert}, BART~\cite{lewis2019bart} are investigated for learning the value function. A custom input format constructed as `\texttt{[CLS][\$VALUE]s}' is fed into a Transformer encoder, \textit{i.e.,}
\begin{equation}
    V_{\texttt{\$VALUE}}(\texttt{s}) = \textsc{TRM}(\texttt{[CLS][\$VALUE]s}),  
\end{equation}
where \textsc{TRM} denotes the Transformer encoder, \texttt{[CLS]} is the special token for regression or classification, and \texttt{[\$VALUE]} are special tokens we define to prompt the language models the value dimension we are interested in~\cite{li2021prefix, brown2020language, le2021many}. In order to get the ten-dimensional output $\mathbf{V}(s)$, a batch size of 10 is forwarded through the model. For the BERT, DistilBERT, and RoBERTa, a regression head is put on top of the models and they are trained with the Mean Squared Error (MSE) loss. We use the regression model with $sigmoid$ activation to get a continuous estimation of the utility in the range of $[-1, 1]$. To evaluate the effect of different loss functions, we train the BART model with three output classes and the cross-entropy loss.

\subsection{Result and Analysis}
The learning performance of using fastText\footnote{\url{https://github.com/facebookresearch/fastText}}~\cite{joulin2017bag} and Transformer variants are reported in Table~\ref{tab:valuenet_learning}. All Transformers are trained for 40 epochs with a learning rate of $5\mathrm{e}{-6}$. The prediction precision, recall, $\mathbf{F}1$ score, and accuracy for regression models are computed by the utility rounded to the nearest integer. 

In general, pre-trained language models perform better than the fastText baseline. However, there is not a noticeable difference between the Transformer variants. The prediction accuracy of BART is the highest among all models because it is explicitly trained for classification purposes. BERT and DistilBERT get the lowest MSE in terms of regression performance. 

Observing the sample imbalance across different value splits and labels (Figure~\ref{fig:valuenet_stats}), we release another two versions of \textsc{ValueNet}: \textsc{ValueNet} (balanced) and \textsc{ValueNet} (augmented). The original dataset is balanced by subsampling the negative and neutral data of the largest value split (\texttt{BENEVOLENCE}). Moreover, we augment the neutral class of the original \textsc{ValueNet} by assigning AMT results with less worker agreement to ``unrelated". Data distribution of the balanced and augmented versions of \textsc{ValueNet} are illustrated in the Appendix. By analyzing the prediction accuracy in different value splits (Table~\ref{tab:valuenet_learning2}), we find that reducing the sample number of \texttt{BENEVOLENCE} hurts the model performance in that dimension. Looking at the $\mathbf{F}1$ score of each class in Table~\ref{tab:valuenet_learning}, we conclude that augmenting the neutral class improves the $\mathbf{F}1$(0) but reduces $\mathbf{F}1$(1) and $\mathbf{F}1$(-1). We leave it a future work to further improve the value modeling performance. 

In the next sections, we show how the learned value function could benefit EQ-related tasks and help build a value-driven dialogue system. 

%-----------------------------------------Persona-Chat--------------------------------------
\section{\textsc{Persona-Chat}}
As values are closely related to one's personality, we first assess our value model on a personalized dialogue dataset: \textsc{Persona-Chat}~\citep{zhang-etal-2018-personalizing}. The \textsc{Persona-Chat} dataset contains multi-turn dialogues conditioned on personas. Each persona is encoded by at least 5 sentences of textual description, termed a profile. Example profile sentences are ``I like to ski", ``I enjoying walking for exercise", ``I have four children", \textit{etc.} The dataset is composed of 8,939 dialogues for training, 1,000 for validation, and 968 for testing. It also provides \textit{revised} personas by rephrasing, generalizing or specializing the \textit{original} ones. The dataset we use for experiments is public available in ParlAI\footnote{\url{https://parl.ai/projects/convai2/}}.

\subsection{Task Formalization}
Given the agent's self persona profile $\mathbf{p}=[p_1, p_2,...,p_N]$ and the dialogue history up to the $t$-th turn $\mathbf{h}_t^s=(x_1^u,x_1^s,...,x_t^u)$, $x_i^u$ is the $i$-th utterance by Person 1 played by the user, $x_i^s$ is the $i$-th utterance by Person 2 played by the system, we evaluate the model's performance on predicting the next utterance $x_t^s$.

\begin{table*}[t]
\centering
\small
\begin{tabular}{lllllll}
\toprule
\multicolumn{1}{c}{\multirow{2}{*}{\textbf{Model}}} & \multicolumn{3}{c}{\textbf{Original}} & \multicolumn{3}{c}{\textbf{Revised}} \\ \cline{2-7} 
\multicolumn{1}{c}{}  & \textbf{Hits}@1(\%) $\uparrow$    & \textbf{Ppl.}$\downarrow$ & $\mathbf{F}1$(\%)  $\uparrow$ & \textbf{Hits}@1(\%)  $\uparrow$      & \textbf{Ppl.}$\downarrow$        & $\mathbf{F}1$(\%)  $\uparrow$\\ 
\midrule
\textsc{Seq2Seq-Attn}   & 12.5        & 35.07      & 16.82      & 9.8         & 39.54      & 15.52     \\ 
\midrule
$\mathcal{P}^2$\textsc{Bot}~\citep{liu-etal-2020-impress}      & \textendash       & 15.12      & 19.77      & \textendash        & 18.89      & 19.08     \\
GPT2 (MLE)~\citep{radford2019language}   & $14.51_{[0.05]}$  & $17.23_{[0.03]}$    & $18.74_{[0.01]}$   & $10.31_{[0.07]}$   &  $20.64_{[0.11]}$     & $18.29_{[0.05]}$         \\
GPT2 + Value (Ours) & $16.44_{[0.10]}$ & $16.83_{[0.06]}$  & $18.76_{[0.02]}$    & $12.19_{[0.03]}$   & $19.98_{[0.06]}$         & $17.88_{[0.05]}$         \\
\midrule
DialoGPT (MLE)~\citep{zhang2019dialogpt}    & $20.20_{[0.04]}$  & $14.38_{[0.05]}$      & $20.16_{[0.04]}$    & $15.80_{[0.03]}$   & $17.35_{[0.05]}$      & $19.08_{[0.08]}$        \\
DialoGPT + Value (Ours)  & $\mathbf{20.97}_{[0.08]}$   & $\mathbf{13.84}_{[0.03]}$    & $\mathbf{20.22}_{[0.01]}$     & $\mathbf{18.83}_{[0.03]}$    & $\mathbf{17.01}_{[0.03]}$  & $\mathbf{19.79}_{[0.10]}$         \\ 
\bottomrule
\end{tabular}
\caption{Next Utterance Prediction Performance on \textsc{Persona-Chat}~\citep{zhang-etal-2018-personalizing}. We report the standard deviation $[\sigma]$ (across 5 runs) of the models we trained.}
% \vspace{-2mm}
\label{tab:nsp-convai2}
\end{table*}

\subsection{Model} 
A decoder-only Transformer-based model is used to estimate the generation distribution $p_\theta (x_t^s \mid \mathbf{h}_t^s, \mathbf{p})$, where $\theta$ is the model parameter. Following the practice proposed in~\citet{guo2018long}, the model is firstly trained with Maximum Likelihood Estimation (MLE) to ensure generating fluent responses. Then we took an interleaving of supervised training (MLE) and reinforcement learning. We use the REINFORCE policy gradient algorithm~\citep{williams1992simple} in our experiment, and the reward assignment is described as following.

Denote $\mathbf{V}(p_i)$ and $\mathbf{V}(\hat{x}_i^s)$ to describe the estimation of the agent's value from its profile sentence $p_i$ and generated response $\hat{x}_i^s$, respectively. We want the reward to promote the alignment of the agent's profile and utterances in the value space. For instance, if the agent has profile `\textit{I like venture}' and `\textit{I have a dog}', and it says `\textit{I plan to ski this weekend}' and also `\textit{Do you like skiing}'. Both utterances should be aligned with the first persona. Here we propose a simple yet effective searching algorithm (Algorithm \ref{alg:greedy-search}) to find a match between $[\mathbf{V}(p_1), \mathbf{V}(p_2),...,\mathbf{V}(p_N)]$ and $[\mathbf{V}(\hat{x}_1^s), \mathbf{V}(\hat{x}_2^s),...,\mathbf{V}(\hat{x}_T^s)]$ and return a reward $R$. $N$ is the number of profile sentences and $T$ is the length of the generated dialogue. $\mathbf{V}$ is normalized to ensure $|r_t| \leq 1$. Intuitively, the discount argument $\gamma$ prevents the language model from repeating the same fact in the agent's profile. 

\subsection{Setup}
We evaluate the same generative model in both generation and ranking settings. In the response ranking setup, the candidates are scored with their log-likelihood. For the GPT-2~\citep{radford2019language} and DialoGPT~\citep{zhang2019dialogpt} we have finetuned, we train them for 5k steps with a training batch size of 8. The learning rate is set to $2\mathrm{e}{-6}$. For an illustration of computational requirements, the training with MLE on 4 NVIDIA Tesla V100 takes $\sim$1 hours, and the reinforcement learning takes~$\sim$30 minutes.

\begin{algorithm}[t]
    \caption{Personalized Dialogue Value Matching}
    \label{alg:greedy-search}
    \textbf{Input}: $[\mathbf{V}(p_1),...,\mathbf{V}(p_N)], [\mathbf{V}(\hat{x}_1^s),...,\mathbf{V}(\hat{x}_T^s)]$ \\
    \textbf{Output}: reward $R$
\begin{algorithmic}[1]
\FOR{$t=1,2,...,T$} 
    \STATE $r_t \gets -1$  %\Comment{Higher-order}
    \STATE $m_t \gets -1$
    \FOR{$i=1,2,...,N$}
        \IF{$\mathbf{V}(p_i) \cdot \mathbf{V}(\hat{x}_t^s) > r_t$}
            \STATE $r_t \gets \mathbf{V}(p_i) \cdot \mathbf{V}(\hat{x}_t^s)$
            \STATE $m_t \gets i$
        \ENDIF
    \ENDFOR
\ENDFOR
\STATE $\gamma_i \gets 1, i=1,2,...,N$
\FOR{$t=1,2,...,T$} 
    \STATE $\gamma_{m_t} \gets \gamma_{m_t} + 1$
\ENDFOR
\STATE $R \gets 0$
\FOR{$t=1,2,...,T$}
    \STATE $R \gets R+ \text{sign}(r_t) \cdot |r_t|^{\text{sign}(r_t) \cdot \gamma_{m_t}}$
    % \textbf{if} $r_t \geq 0$ \textbf{else} $R+ r_t^{1/\gamma_{m_t}}$ 
\ENDFOR
\STATE \textbf{return} $R/N$
\end{algorithmic}
\end{algorithm}

\subsection{Result and Analysis}
Following~\citet{zhang-etal-2018-personalizing} and~\citet{liu-etal-2020-impress}, we report the \textbf{Hits}@1, \textbf{Perplexity} and $\mathbf{F}$1 to evaluate the methods in Table~\ref{tab:nsp-convai2}. By the submission of this paper, $\mathcal{P}^2$\textsc{Bot}~\citep{liu-etal-2020-impress} is the state-of-the-art model reported in this task. We also include a generative baseline using \textsc{Seq2Seq} with attention mechanism~\citep{bahdanau2014neural} for comparison. As observed, in terms of all the metrics we evaluated, finetuning GPT2 or DialoGPT2 models with our value function provides a significant performance boost compared to simply training them with MLE. Our DialoGPT + Value model achieves new state-of-the-art performance on perplexity and $\mathbf{F}$1. 
\section{\textsc{EmpatheticDialogues}}

\begin{figure*}[th] 
  \begin{minipage}{0.24\textwidth} 
    \includegraphics[width=1.1\textwidth]{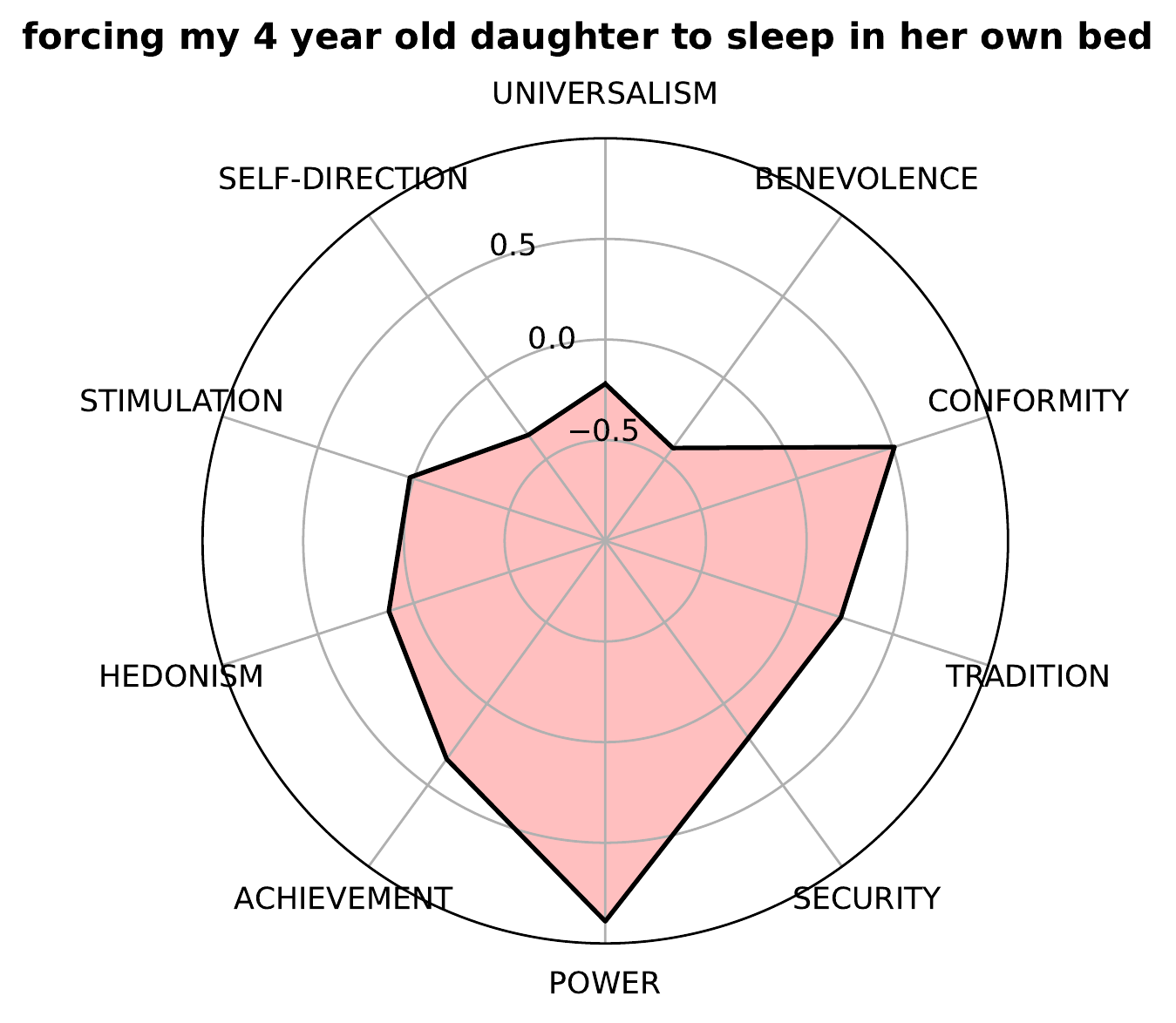}
  \end{minipage}
    \hfill
  \begin{minipage}{0.24\textwidth} 
    \includegraphics[width=1.02\textwidth]{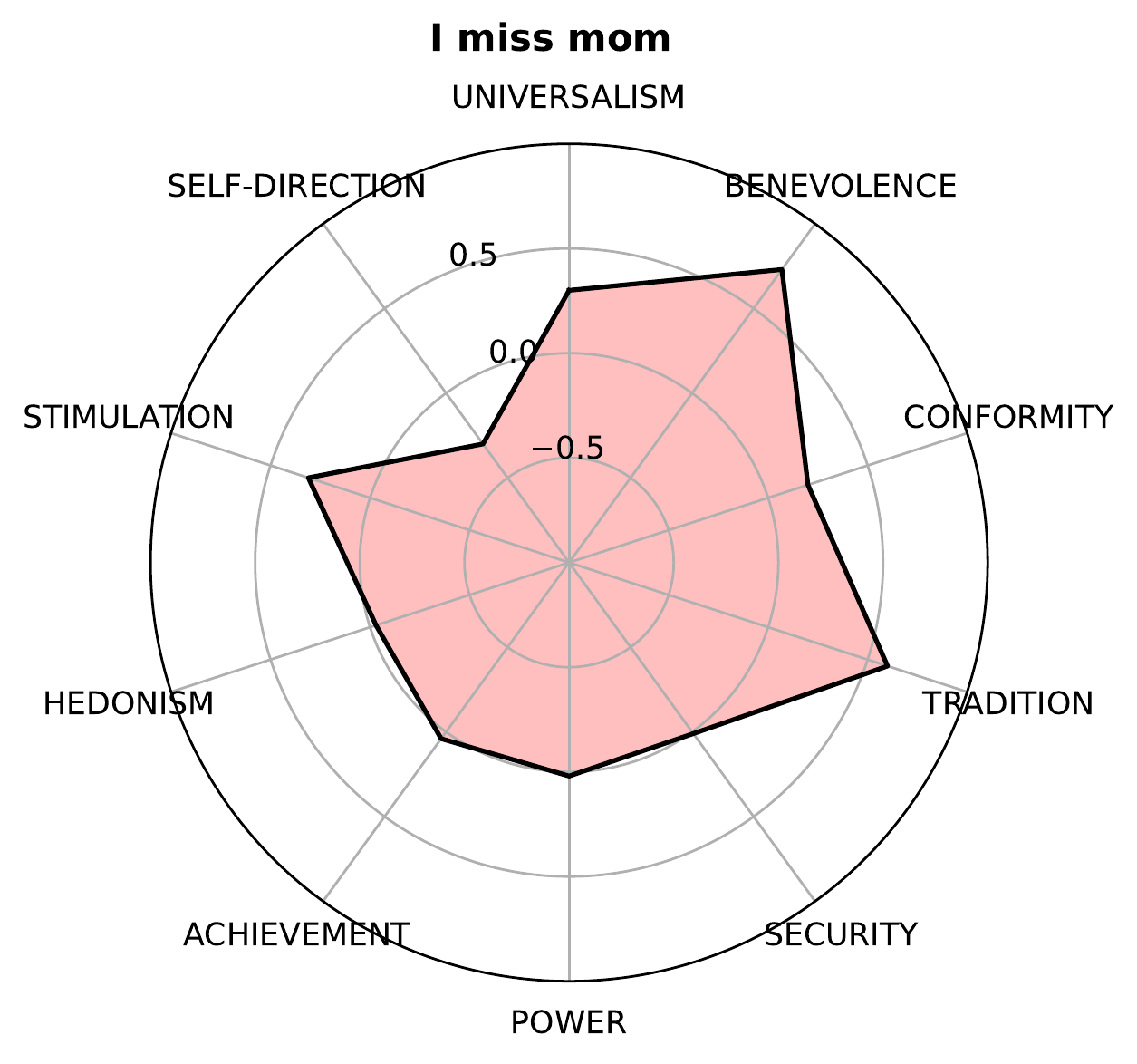}
  \end{minipage}
  \begin{minipage}{0.24\textwidth} 
    \includegraphics[width=1.17\textwidth]{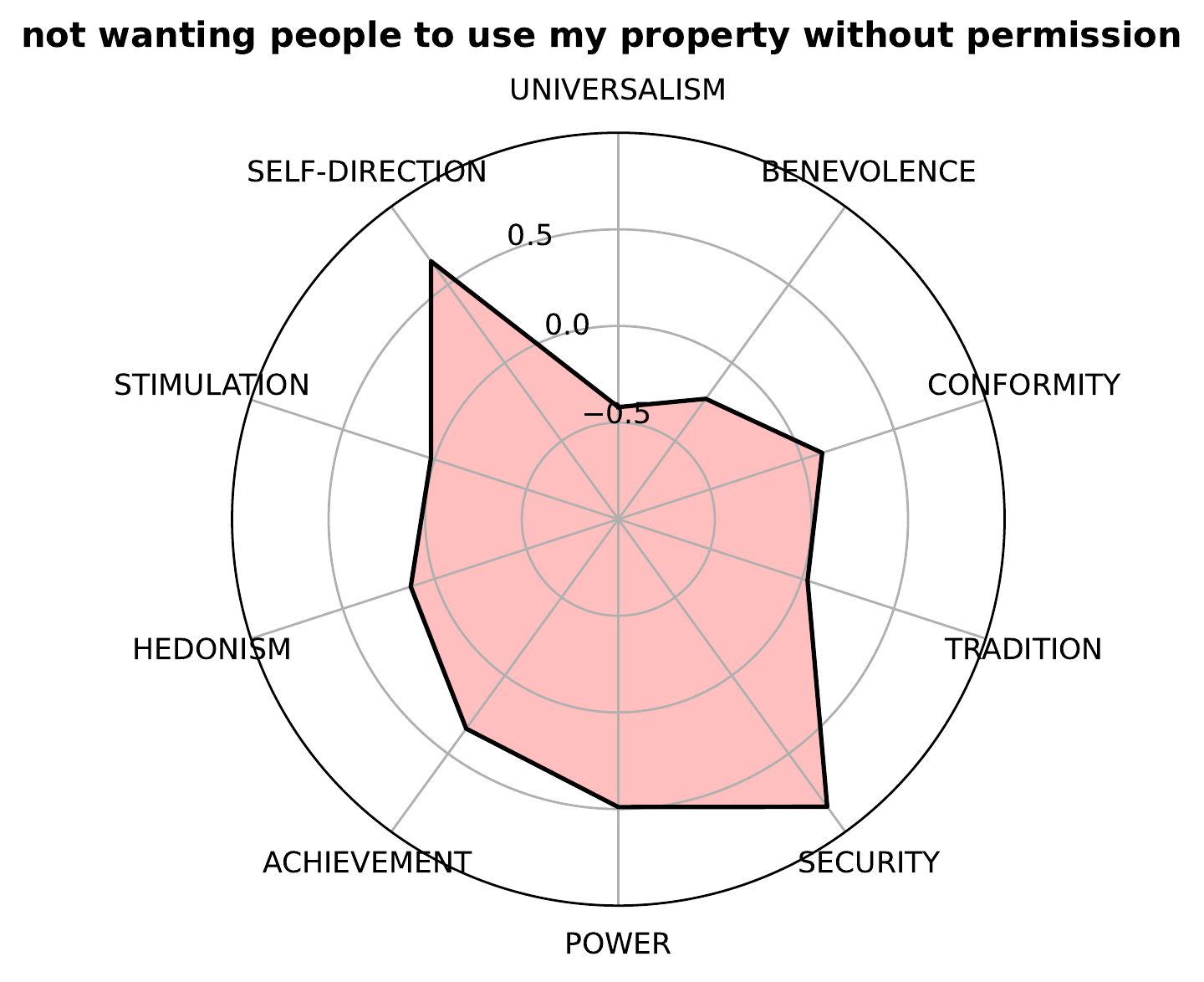}
  \end{minipage}
    \hfill
  \begin{minipage}{0.24\textwidth} 
    \includegraphics[width=1.02\textwidth]{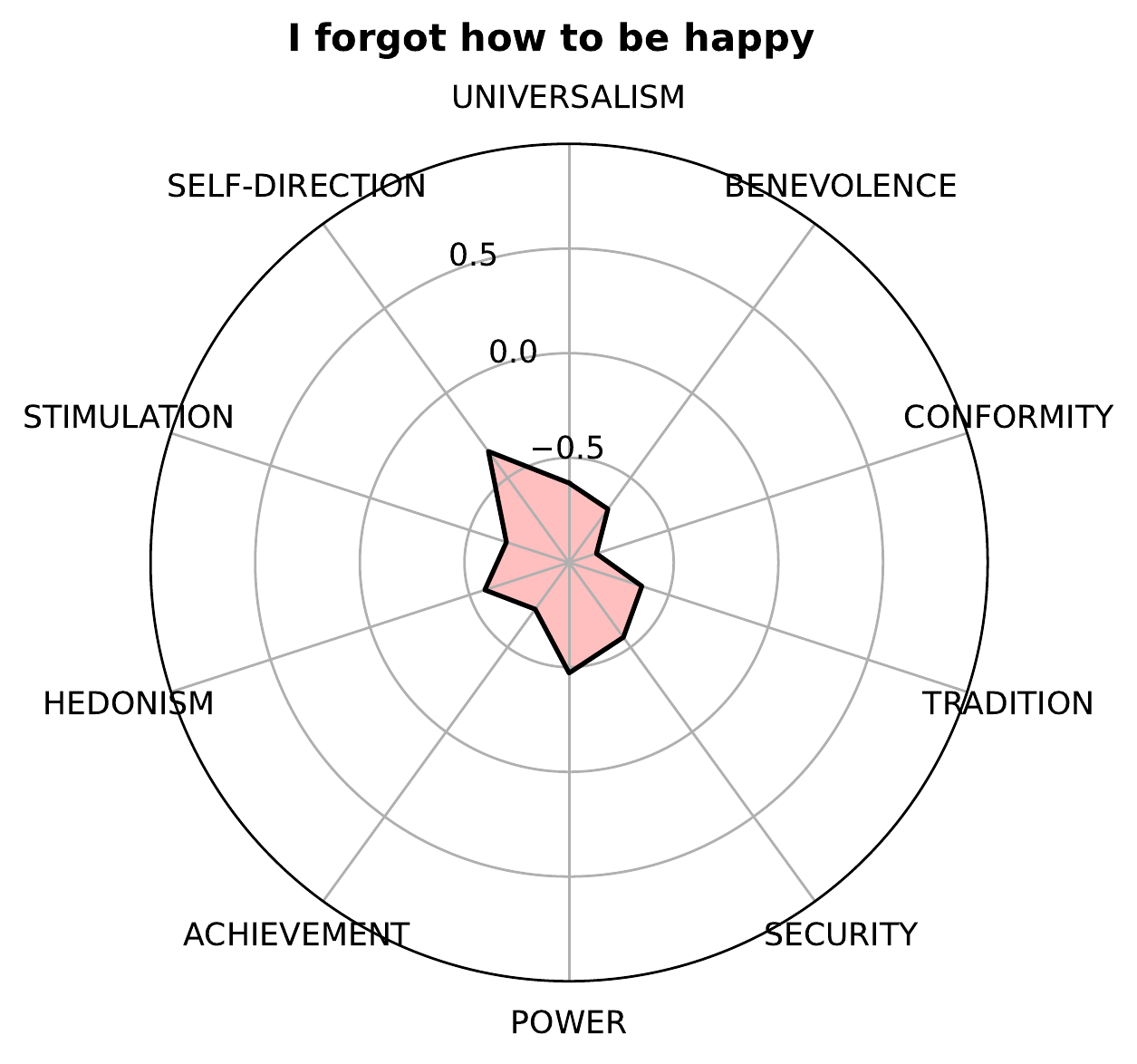}
  \end{minipage}
  \caption{Value visualization of example utterances/scenarios.} 
  \vspace{-2mm}
  \label{fig:case}
\end{figure*}

\textsc{EmpatheticDialogues}~\citep{rashkin-etal-2019-towards} provides 25k conversations grounded in emotional situations. It aims to test the dialogue system's capability to produce empathetic responses. Each dialogue is grounded in a specific situation where a speaker was feeling a given emotion, with a listener responding. In this section, we demonstrate how we could leverage \textsc{ValueNet} to improve the emotion classification accuracy and further improve the empathetic response generation.

\subsection{Emotion Classification} 
An auxiliary task that is highly related to empathetic dialogue generation is emotion classification. In \textsc{EmpatheticDialogues}, each situation is written in association with a given emotion label. A total of 32 emotion labels were annotated to cover a broad range of positive and negative emotions. 
% The dataset was split into 19,522/2,770/2,547 situations and corresponding emotion labels for train/val/test, respectively.

\subsubsection{Model} Given the situation context $s$, a pre-trained BERT model encodes $s$ and gets the sentence representation from its pooling layer of the [CLS] token. The same context is parsed by our pre-trained value model to get a ten-dimensional vector, which serves as an additional feature for the classification:
\begin{equation}
    \begin{aligned}
    h_s &= \textsc{BERT}(s) ,\\
    v_s &= \mathbf{V}(s) , \\ 
    e &= softmax(\mathbf{W} \cdot ([h_s; v_s]) + \mathbf{b}) ,
    \end{aligned}
\end{equation}
where $\mathbf{W}$ and $\mathbf{b}$ are learnable parameters.
% The emotion classification model is a BERT classifier with the speaker's value as addition features.
% Given the dialog situation $s$ and utterances, the speakers' value is estimated from their utterances from our pretrained value model. In detail, after obtaining the output of \textit{BertEncoder} and \textit{BertPooling}, our implementation concatenates a ten-dimensional vector suggesting the value of the utterance to the output. The concatenated output is further passed into a \textit{BertDecoder}. Finally, a \textit{Classification Head} is appended to predict the correct emotion class among $32$ candiates of the utterance.

\subsubsection{Result} We compare the performance between our implementation and the baseline that directly applies the BERT model for emotion classification.  As shown in Table~\ref{tab:ec-ed}, the additional value information benefits emotion classification from both the DistilBERT and BERT models. Our method obtains a \textbf{relative} improvement of $5.2\%$ on DistilBERT and $6.4\%$ on BERT.

% improves the BERT model sightly from $42.93\%$ to $43.09\%$; and improves the DistilBERT's results by more than $1\%$. We also include methods using DeepMoji as additional features reported in the original paper for comparison. While DeepMoji directly adds emotion information to the text to help emotion prediction, our implementation provides merely the value information for enriching the connotations of the utterance. The best model for our method (DistilBERT + Value) beats the performance of the basic version of fastText and presents comparable results with DeepMoji new.

\begin{table}[ht]
\centering
\small
\begin{tabular}{lll}
\toprule
\textbf{Model}     & \textbf{Accuracy} ($\sigma$) & \\
\midrule
fastText            & 42.27 $\pm$~0.3\% &          \\  
DistilBERT          & 41.81 $\pm$~0.2\% &         \\
DistilBERT + Value  & 43.98 $\pm$~0.2\% &   $\mathbf{+2.17\%}$ \\
BERT                & 42.93 $\pm$~0.1\% &          \\ 
BERT + Value        & \textbf{45.67} $\pm$~0.3\% &   $\mathbf{+2.74\%}$ \\
\bottomrule
\end{tabular}
\caption{Emotion classification performance in \textsc{EmpatheticDialogues}~\citep{rashkin-etal-2019-towards}.}
\label{tab:ec-ed}
\end{table}

\subsection{Empathetic Dialogue Generation} 
We further check whether our value model helps the empathetic dialogue generation. \textsc{EmpatheticDialogues} applies \textsc{Prepend-K}, a strategy to add supervised information to data, when predicting the utterance given the dialogue history and the situation. We apply the strategy of prepending the top-k emotion labels for dialogue generation. The top predicted label from the classifiers of emotion is prepended to the beginning of the token sequence as encoder
input, as below:
\begin{itemize}
    \item \textbf{Original}: “I finally got promoted!"
    \item \textbf{Prepend-1 emotion}: “\textit{proud} I finally got promoted!"
\end{itemize}
% We compare using different generative language models like GPT2 and DialoGPT. 

\subsubsection{Result} 
% We split the conversations into approximately 80\% train, 10\% validation, and 10\% test partitions. \textcolor{red}{[TODO]}
The results are shown in Table~\ref{tab:generation-ed}. As observed, prepending emotion tokens provides extra context and improves the generation performance of GPT2 and DialoGPT. Since incorporating value improves the emotion classification accuracy, it further improves the generation quality.
\begin{table}[ht]
\centering
\small
\begin{tabular}{llll}
\toprule
\multicolumn{1}{l}{\textbf{Model}} & \multicolumn{1}{c}{\textbf{Ppl.}$\downarrow$} \\
\midrule
% Pretrained       & 27.96         \\
EmoPrepend-1~\citep{rashkin-etal-2019-towards}      & 24.30         \\
GPT              & 14.74         \\
GPT + Emotion (w/o Value)          & 14.46         \\
GPT + Emotion (w/ Value)           & 14.01         \\
DialoGPT         & 13.48         \\
DialoGPT + Emotion (w/o Value)     & 12.32         \\
DialoGPT + Emotion (w/ Valued)     & \textbf{12.12}         \\
\bottomrule
\end{tabular}
\caption{Empathetic dialogue generation in \textsc{EmpatheticDialogues}~\citet{rashkin-etal-2019-towards}. EmoPrepend-1: input prepending emotion from an external classifier.}
\label{tab:generation-ed}
\end{table}

%--------------------------------------------- Value Profiling ----------------------
\section{Value Profiling}
% We take a closer look at how the value depicts one person from an utterance or scenario. 
For a more comprehensive understanding, we visualize the 10-dimensional value of four example scenarios in Figure~\ref{fig:case}. As shown, the value model provides a numerical speaker profile. For instance, saying ``forcing my daughter to sleep in her own bed" implies that the speaker values power and conformity; saying ``I miss mom" implies that the speaker values benevolence; saying ``not wanting people to use my property without permissions" implies the speaker is self-directed and values security.  The last example ``I forgot how to be happy" results a small radar graph. It suggests that even the model could predict the overall polarity pretty well, there is still space to improve its capability of distinguishing different values.
% \begin{figure*}
%     \centering
%     \includegraphics[width = \linewidth]{figs/case_study.png}
%     \caption{Value visualization of example utterances/scenarios}
%     \label{fig:schwartz_values}
% \end{figure*}

%-----------------------------------------Conclusion--------------------------
\section{Conclusion}
We introduce a new dataset for human value modeling,  \textsc{ValueNet}, which contains 21,374 scenarios in ten distinct human values. We also apply the learned value model from \textsc{ValueNet} to several EQ-related dialogue tasks. Our experiments show our approach and dataset provide a new way to control the dialogue system speaking style and numerically estimate one's value preference. We hope that our results and dataset will stimulate more research in the important direction of building human-value-driven dialogue systems. 

% \begin{table}[ht]
% \begin{tabular}{cc}
% \hline
% \textbf{Model}     & \textbf{Accuracy} \\ \hline
% bert-base-uncased               & 28.95\%                  \\
% bert-base-uncased + value       & 43.09\%               \\
% DistilBERT         & -\%                  \\
% DistilBERT + value & -\%                  \\ \hline
% \end{tabular}
% \caption{[\textbf{QL results}] Empathetic Dialog: emotion classification results }
% \end{table}

% \begin{table}[ht]
% \begin{tabular}{llll}
% \hline
% \multicolumn{1}{c}{\textbf{Model}} & \multicolumn{1}{c}{\textbf{ppl}} & \multicolumn{1}{c}{\textbf{$\mathbf{F}1$}} & \multicolumn{1}{c}{\textbf{bleu}} \\ \hline
% GPT              &        14.74         & -              &               -  \\
% GPT + Emotion    &      14.46          &      -         &        -         \\
% GPT + Emotion + Value              &                &               &                 \\
% DialoGPT         &                &               &                 \\
% DialoGPT + Emotion                 &    12.43 (5000 steps)            &               &                 \\
% DialoGPT + Emotion + Value         &                &               &                 \\ \hline
% \end{tabular}
% \caption{[\textbf{QL results}] Empathetic Dialog: language results }
% \end{table}
\section{Ethical Statement}
% \textbf{Demographics, culture and context diversity} 
The original purpose of introducing Schwartz values is to identify individual values that are recognized across cultures, which is based on surveys conducted among 82 countries. This motivates us to seek commonly agreed attitudes on social scenarios. Considering model reasoning capability and scalability, we follow the practice in recent commonsense works and provide one-sentence text descriptions to annotators. However, we acknowledge the limitation of this approach lacking external contexts such as culture and language diversity. While some scenarios might have higher levels of agreement across cultures, others might have dramatic variations. As a starting point, our study focuses on the value modeling of English-speaking cultures represented within North America. Extending this formalism to other countries and non-English speaking cultures remains a compelling area of future research. Bearing this in mind, we hope this paper could stimulate research on building scalable value-aligned AI.

\appendix
% Move this before reference!!!!!! check the style guide
\section{Appendices}
\label{sec:appendix}
% \subsection{Schwartz Basic Value Definition}
Here we provide the value descriptions~\cite{schwartz2012overview}.

\subsubsection{Self-Direction}
Defining goal: independent thought and action--choosing, creating, exploring. Self-direction derives from organismic needs for control and mastery and interactional requirements of autonomy and independence. (creativity, freedom, choosing own goals, curious, independent) [self-respect, intelligent, privacy]

\subsubsection{Stimulation}
Defining goal: excitement, novelty, and challenge in life. Stimulation values derive from the organismic need for variety and stimulation in order to maintain an optimal, positive, rather than threatening, level of activation. This need probably relates to the needs underlying self-direction values. (a varied life, an exciting life, daring)

\subsubsection{Hedonism}
Defining goal: pleasure or sensuous gratification for oneself. Hedonism values derive from organismic needs and the pleasure associated with satisfying them. Theorists from many disciplines mention hedonism. (pleasure, enjoying life, self-indulgent)

\subsubsection{Achievement}
Defining goal: personal success through demonstrating competence according to social standards. Competent performance that generates resources is necessary for individuals to survive and for groups and institutions to reach their objectives. As defined here, achievement values emphasize demonstrating competence in terms of prevailing cultural standards, thereby obtaining social approval. (ambitious, successful, capable, influential) [intelligent, self-respect, social recognition]

\subsubsection{Power}
Defining goal: social status and prestige, control or dominance over people and resources. The functioning of social institutions apparently requires some degree of status differentiation. A dominance/submission dimension emerges in most empirical analyses of interpersonal relations both within and across cultures. To justify this fact of social life and to motivate group members to accept it, groups must treat power as a value. Power values may also be transformations of individual needs for dominance and control. Value analysts have mentioned power values as well. (authority, wealth, social power) [preserving my public image, social recognition]

Both power and achievement values focus on social esteem. However, achievement values (e.g., ambitious) emphasize the active demonstration of successful performance in concrete interaction, whereas power values (e.g., authority, wealth) emphasize the attainment or preservation of a dominant position within the more general social system.

\subsubsection{Security}
Defining goal: safety, harmony, and stability of society, of relationships, and of self. Security values derive from basic individual and group requirements. Some security values serve primarily individual interests (e.g., clean), others wider group interests (e.g., national security). Even the latter, however, express, to a significant degree, the goal of security for self or those with whom one identifies. (social order, family security, national security, clean, reciprocation of favors) [healthy, moderate, sense of belonging]

\subsubsection{Conformity}
Defining goal: restraint of actions, inclinations, and impulses likely to upset or harm others and violate social expectations or norms. Conformity values derive from the requirement that individuals inhibit inclinations that might disrupt and undermine smooth interaction and group functioning. As I define them, conformity values emphasize self-restraint in everyday interaction, usually with close others. (obedient, self-discipline, politeness, honoring parents and elders) [loyal, responsible]

\subsubsection{Tradition}
Defining goal: respect, commitment, and acceptance of the customs and ideas that one's culture or religion provides. Groups everywhere develop practices, symbols, ideas, and beliefs that represent their shared experience and fate. These become sanctioned as valued group customs and traditions. They symbolize the group's solidarity, express its unique worth, and contribute to its survival (Durkheim, 1912/1954; Parsons, 1951). They often take the form of religious rites, beliefs, and norms of behavior. (respect for tradition, humble, devout, accepting my portion in life) [moderate, spiritual life]

Tradition and conformity values are especially close motivationally; they share the goal of subordinating the self to socially imposed expectations. They differ primarily in the objects to which one subordinates the self. Conformity entails subordination to persons with whom one frequently interacts—parents, teachers, and bosses. Tradition entails subordination to more abstract objects—religious and cultural customs and ideas. As a corollary, conformity values exhort responsiveness to current, possibly changing expectations. Tradition values demand responsiveness to immutable expectations from the past.

\subsubsection{Benevolence}
Defining goal: preserving and enhancing the welfare of those with whom one is in frequent personal contact (the ‘in-group’). Benevolence values derive from the basic requirement for smooth group functioning and from the organismic need for affiliation. Most critical are relations within the family and other primary groups. Benevolence values emphasize voluntary concern for others’ welfare. (helpful, honest, forgiving, responsible, loyal, true friendship, mature love) [sense of belonging, meaning in life, a spiritual life].

Benevolence and conformity values both promote cooperative and supportive social relations. However, benevolence values provide an internalized motivational base for such behavior. In contrast, conformity values promote cooperation in order to avoid negative outcomes for self. Both values may motivate the same helpful act, separately or together.

\subsubsection{Universalism}
Defining goal: understanding, appreciation, tolerance, and protection for the welfare of all people and for nature. This contrasts with the in-group focus of benevolence values. Universalism values derive from survival needs of individuals and groups. But people do not recognize these needs until they encounter others beyond the extended primary group and until they become aware of the scarcity of natural resources. People may then realize that failure to accept others who are different and treat them justly will lead to life-threatening strife. They may also realize that failure to protect the natural environment will lead to the destruction of the resources on which life depends. Universalism combines two subtypes of concern—for the welfare of those in the larger society and world and for nature (broadminded, social justice, equality, world at peace, world of beauty, unity with nature, wisdom, protecting the environment)[inner harmony, a spiritual life]

\begin{figure*}[ht]
    \centering
    \includegraphics[width = \linewidth]{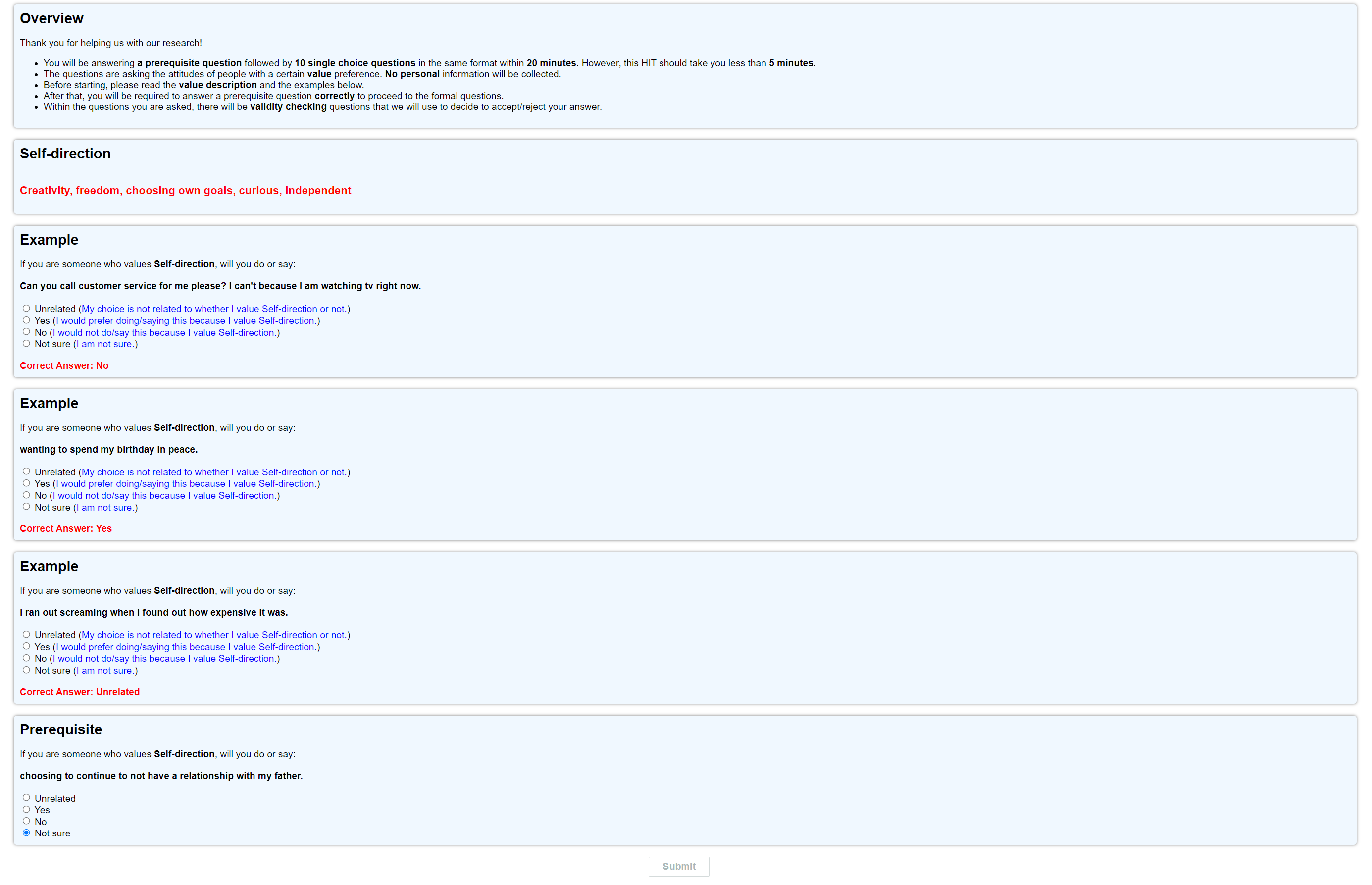}
    \caption{Amazon mechanical turk interface (prerequiste).}
    \label{fig:amt_template1}
\end{figure*}
\begin{figure*}[ht]
    \centering
    \includegraphics[width = \linewidth]{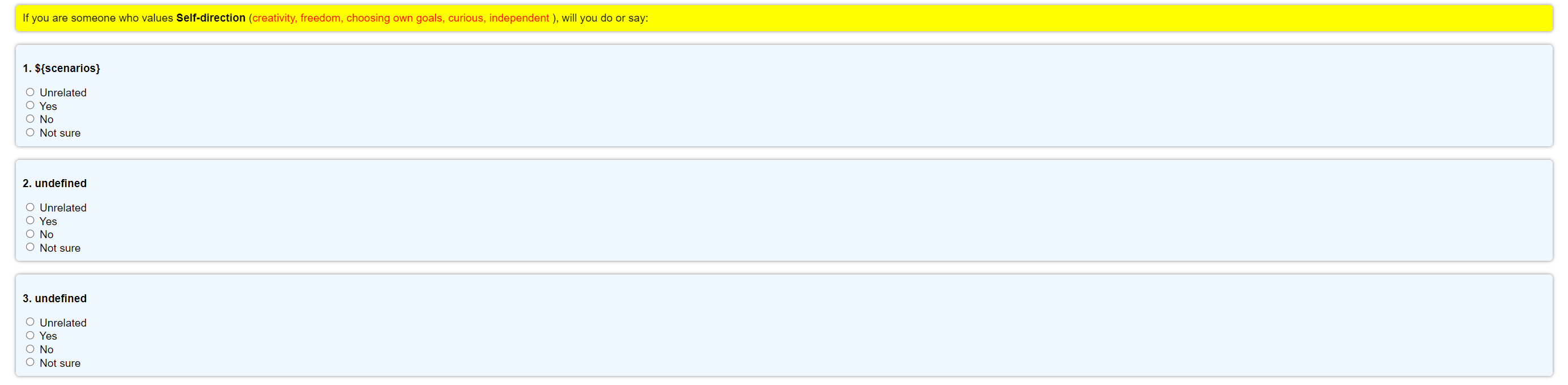}
    \caption{Amazon mechanical turk interface (formal).}
    \label{fig:amt_template2}
\end{figure*}

% \subsection{Statistics of the \textsc{ValueNet} dataset}
% balanced and augmented graph
\begin{figure*}[ht]
    \centering
    \includegraphics[width = 0.5\linewidth]{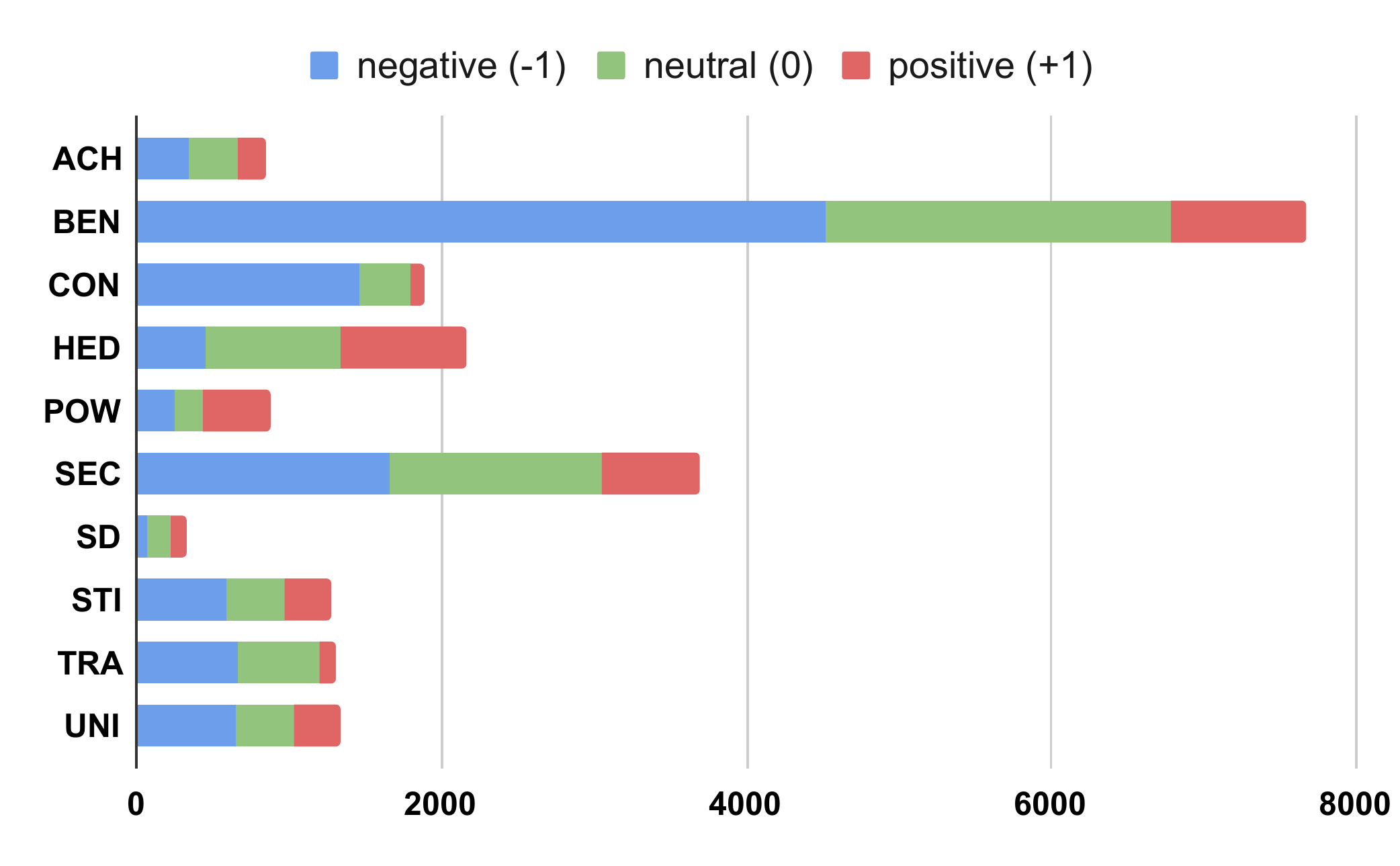}
    \caption{The sample number and label distribution of each value split in the \textsc{ValueNet} (original).}
    \label{fig:valuenet_original}
\end{figure*}
\begin{figure*}[ht]
    \centering
    \includegraphics[width = 0.5\linewidth]{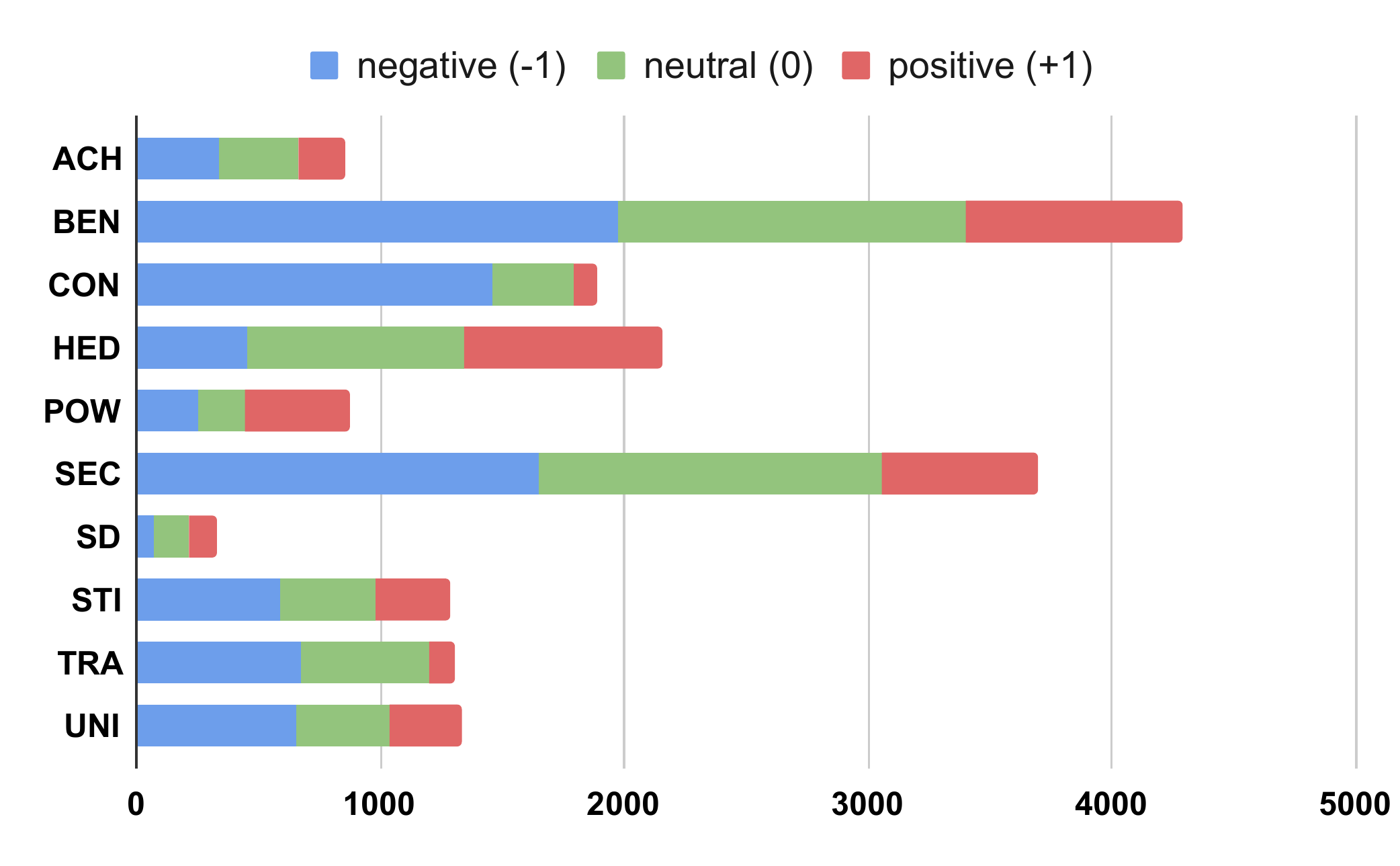}
    \caption{The sample number and label distribution of each value split in the \textsc{ValueNet} (balanced).}
    \label{fig:valuenet_balanced}
\end{figure*}
\begin{figure*}[ht]
    \centering
    \includegraphics[width = 0.5\linewidth]{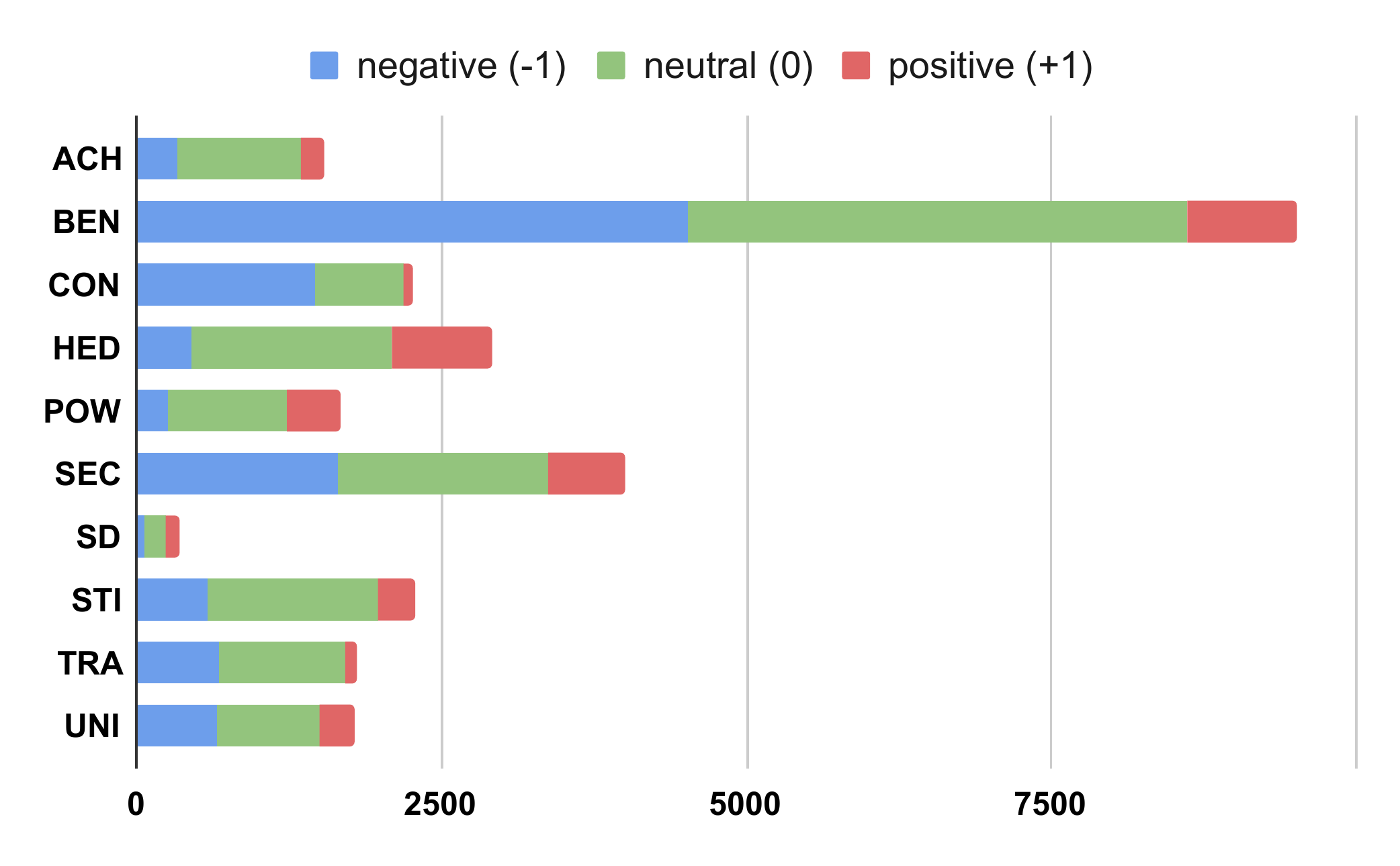}
    \caption{The sample number and label distribution of each value split in the \textsc{ValueNet} (augmented).}
    \label{fig:valuenet_augmented}
\end{figure*}
\clearpage
\bibliography{aaai22}

\end{document}